\documentclass[letterpaper, 10 pt, conference]{ieeeconf}  
\IEEEoverridecommandlockouts                              %

\usepackage{verbatim}
\usepackage{amssymb}
\usepackage{mathtools}
\usepackage{commath}
\usepackage{nth}
\usepackage{cite}
\usepackage{cleveref}
\usepackage{caption,url}
\usepackage{subcaption}
\usepackage{xcolor}
\usepackage{dblfloatfix} 
\usepackage{float} 

\usepackage{soul}

\usepackage{version}

\excludeversion{old_text}

\usepackage[vlined,ruled]{algorithm2e}

\usepackage {tikz}
\usetikzlibrary{shapes ,arrows , positioning }
\def\prob{\mathbb{P}}

\usepackage{amsmath,graphicx,epsfig,color,amsfonts}

\graphicspath{{./fig/}}

\def\prob{\mathbb{P}}

\newcommand{\seqdef}[2]{\{#1\}_{#2}}

\newcommand\oprocendsymbol{\hbox{$\square$}}
\newcommand\oprocend{\relax\ifmmode\else\unskip\hfill\fi\oprocendsymbol}

\def \mc {\mathcal}

\renewcommand{\theenumi}{(\roman{enumi}}

\usepackage{caption}
\renewcommand{\baselinestretch}{0.98}

\title{Modeling Trust Dynamics in Robot-Assisted Delivery: \\
Impact of Trust Repair Strategies
\thanks{This work was supported in part  by the NSF award ECCS 2024649 and the ONR
award N00014-22-1-2813.}
}

\author{Dong Hae Mangalindan, Karthik Kandikonda, Ericka Rovira, and Vaibhav Srivastava
\thanks{D. Mangalindan and V. Srivastava are with the Department of Electrical and Computer Engineering, Michigan State University, East Lansing, MI 48824  e-mail: \texttt{\{mangalin, vaibhav\}@egr.msu.edu}}
\thanks{K. Kandikonda is a Computer Science Graduate from Michigan State University e-mail: \texttt{kandikon@msu.edu}}
\thanks{E. Rovira is with the Department of Engineering Psychology, U.S. Military Academy, West Point, NY 10996  e-mail: \texttt{ericka.rovira@westpoint.edu}}
}

\begin{document}

\maketitle

\begin{abstract}
With increasing efficiency and reliability, autonomous systems are becoming valuable assistants to humans in various tasks. In the context of robot-assisted delivery, we investigate how robot performance and trust repair strategies impact human trust.
In this task,  while handling a secondary task, humans can choose to either send the robot to deliver autonomously or manually control it.
The trust repair strategies examined include short and long explanations, apology and promise, and denial.
Using data from human participants, we model human behavior using an Input-Output Hidden Markov Model (IOHMM) to capture the dynamics of trust and human action probabilities. Our findings indicate that humans are more likely to deploy the robot autonomously when their trust is high. Furthermore, state transition estimates show that long explanations are the most effective at repairing trust following a failure, while denial is most effective at preventing trust loss.
We also demonstrate that the trust estimates generated by our model are isomorphic to self-reported trust values, making them interpretable. This model lays the groundwork for developing optimal policies that facilitate real-time adjustment of human trust in autonomous systems.
\end{abstract}

\IEEEpeerreviewmaketitle

\section{Introduction}

With the advancement of autonomous robots, their collaboration with human workers becomes crucial to improving productivity and efficiency. Their presence in various sectors is becoming increasingly common, aiming to assist and reduce the workload for humans in various tasks \cite{keshvarparast2024collaborative, silvera2024robotics, gupta2024structural}.
A critical factor in the success of the human-robot collaboration is trust, which is defined by Lee and See \cite{lee2004trust}  as ``the attitude that an agent will help achieve an individual’s goals in a situation characterized by uncertainty and vulnerability''

For effective human-robot teaming, it is essential to appropriately calibrate trust based on robot- and environment-related factors such as capability, complexity, and risk\cite{lee2004trust, parasuraman1997humans}. 
Trust has been extensively studied in both human-human and human-automation interactions. In the former, trust is shaped by mutual risk and reliance, whereas in the latter, it largely depends on the system's performance. Unlike traditional automation, modern robotic systems operate in close proximity to humans, positioning human-robot teams in an intermediate space between human-human and human-automation teams. This raises interesting questions regarding the effectiveness of trust repair strategies---such as apology, promise, denial, and explanation---that are commonly used in human-human teams when applied to human-robot interactions~\cite{baker2018toward}.

In this study, we focus on robot-assisted delivery tasks, investigating how different robot strategies—such as apologies, promises, explanations, and denials—affect human trust following a failure.
To achieve this, we model human behavior using an Input-Output Hidden Markov Model (IOHMM) and estimate the model parameters from experimental data with human participants. By modeling human trust, we can estimate the hidden trust state in real-time, allowing for its calibration accordingly.
We consider a hospital setting where humans work alongside robots to deliver medicine to patients. In this scenario, robots can either autonomously deliver or be manually controlled by the human. This setup reflects real-world collaborative environments, such as warehouses, factories, and construction sites, where robots support humans by taking over time-consuming or repetitive tasks, allowing humans to focus on more complex, decision-intensive responsibilities.

Previous research has modeled human trust using various frameworks, such as linear dynamic models \cite{akash2017dynamic, azevedo2021real, mangalin2024} and Bayesian/POMDP frameworks \cite{OPTIMO, liao2024trust, chen2020trust, zahedi2023trust, DHM-ER-VS:22e,bhat2022clustering,bhat2024evaluating,mangalindan2024trustawareassistanceseekinghumansupervised}; a survey is presented in~\cite{wang2023human}. 
Moreover, studies have explored the impact of role hierarchy and trust repair strategies like promise and explanation on human trust \cite{karli2023if}. In addition to explanation and promise, studies have also explored the effects of other strategies, like denial and apology, on human trust\cite{esterwood2023three, esterwood2023theory,marinaccio2015framework,robinette2015timing}. The trust repair strategies can be grouped into two categories: central and peripheral strategies. Central repair strategies provide more detailed and relevant information, while peripheral repair strategies offer simpler and less relevant information, allowing for quicker and easier processing by the human. The effectiveness of trust repair strategies in relation to the type of autonomous system, the situational context, and human characteristics has been studied \cite{pak2024theoretical}. 

As compared to existing work, our work differs by focusing on modeling human trust, which allows for real-time estimation and investigating how different trust repair strategies---apology and promise, short explanation, long explanation, and denial---impact human trust after a failure. 
By modeling and understanding these dynamics, we can inform the design of more reliable and trustworthy robotic systems that enhance collaborative efficiency. To the best of our knowledge, this is the first computational model to specifically address trust repair strategies.

The major contributions of this work are threefold. First, we design and conduct a human-robot collaborative medicine delivery experiment to assess the effectiveness of the above trust repair strategies. Second, we employ an IOHMM framework to develop a computational model of trust dynamics, with robot performance and trust repair communication as key inputs. Third, we compare the trust estimates generated by our model with self-reported trust values and show that our hidden trust state is interpretable.

The remainder of the paper is structured as follows. Section~\ref{sec:int_task} describes the medicine delivery task and outlines the experimental procedure. In Section~\ref{sec:IOHMMMODEL}, we introduce the IOHMM framework used to model trust dynamics, followed by a discussion of the estimated trust-based human behavior model in Section~\ref{sec:Model}. We compare our estimates of trust with the self-reported trust values from human participants in Section~\ref{sec:grounding}, and finally, we conclude in Section~\ref{sec:conclusion}.

\section{Robot-assisted Delivery Experiment}\label{sec:int_task}

In this section, we describe the collaborative human-robot medicine delivery experiment and the procedure. 

\subsection{Experiment Description}
Consider a scenario in a hospital where a human worker is tasked with counting and sorting medications for delivery to patients' rooms. To assist with this, a fleet of autonomous robots is deployed, capable of autonomously handling the medication deliveries. In our experiment, the human can choose to either allow an autonomous robot to make the delivery or tele-operate the robot. In this scenario, humans and robots collaborate to deliver medications to patients' rooms, while the human also engages in a concurrent task of counting medications. 

\begin{figure}[ht!]
    \centering
    \fbox{
        \includegraphics[width=0.9\linewidth, trim = 10 0 10 30 ]{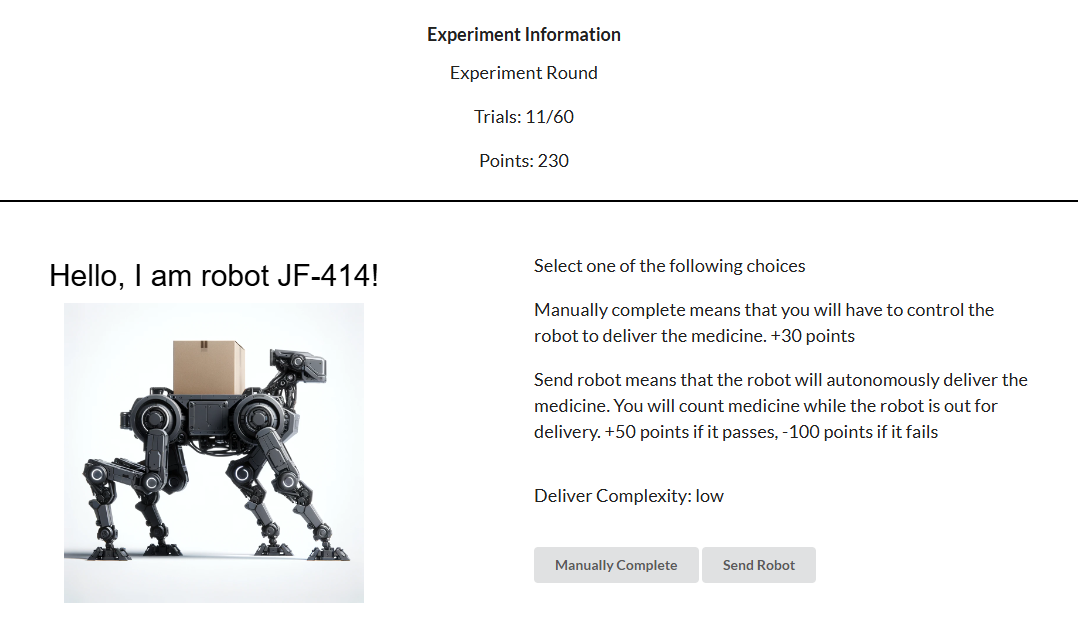}}
        \caption{The beginning of a trial where the robot offers assistance and the human chooses their action as they are informed of the delivery complexity.}
        \label{fig:choice}
    \end{figure}

Each trial consists of a delivery attempt and a medicine counting task. The complexity of the delivery for each trial can be either low ($C^L$) or high ($C^H$), depending on whether the medication needs to be delivered to a single room or multiple rooms. At the start of each trial, a robot from the fleet offering assistance to the human is introduced and the human decides whether to let the robot do the delivery autonomously ($a^{H+}$) or perform the delivery manually ($a^{H-}$). The interface is shown in Fig.~\ref{fig:choice}.

\begin{figure}[ht!]
    \centering
    \fbox{
        \includegraphics[width=0.9\linewidth, trim = 10 20 10 20 ]{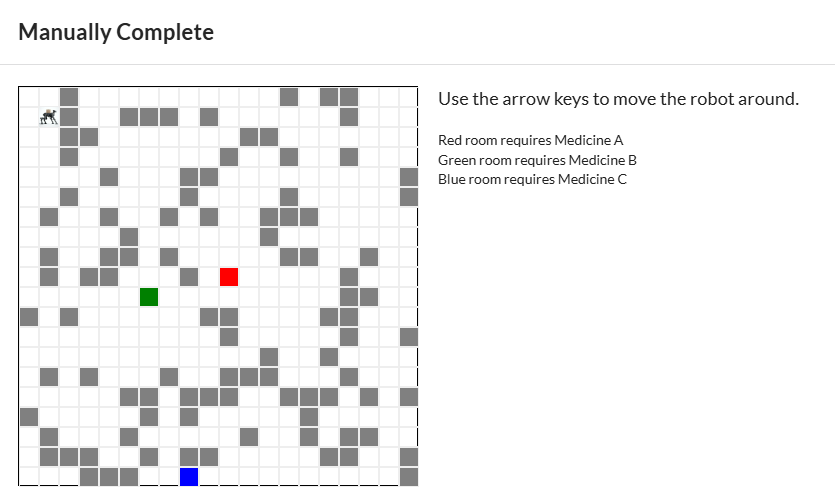}}
        \caption{The interface when a human chooses to manually deliver. They tele-operate the robot and deliver a set of medications to the respective rooms.}
        \label{fig:manua}
    \end{figure}

Manual delivery involves the human tele-operating the robot to the designated locations. The interface for manual delivery is shown in Fig.~\ref{fig:manua}.

\begin{figure}[ht!]
    \centering
    \fbox{
        \includegraphics[width=0.9\linewidth, trim = 10 20 10 30 ]{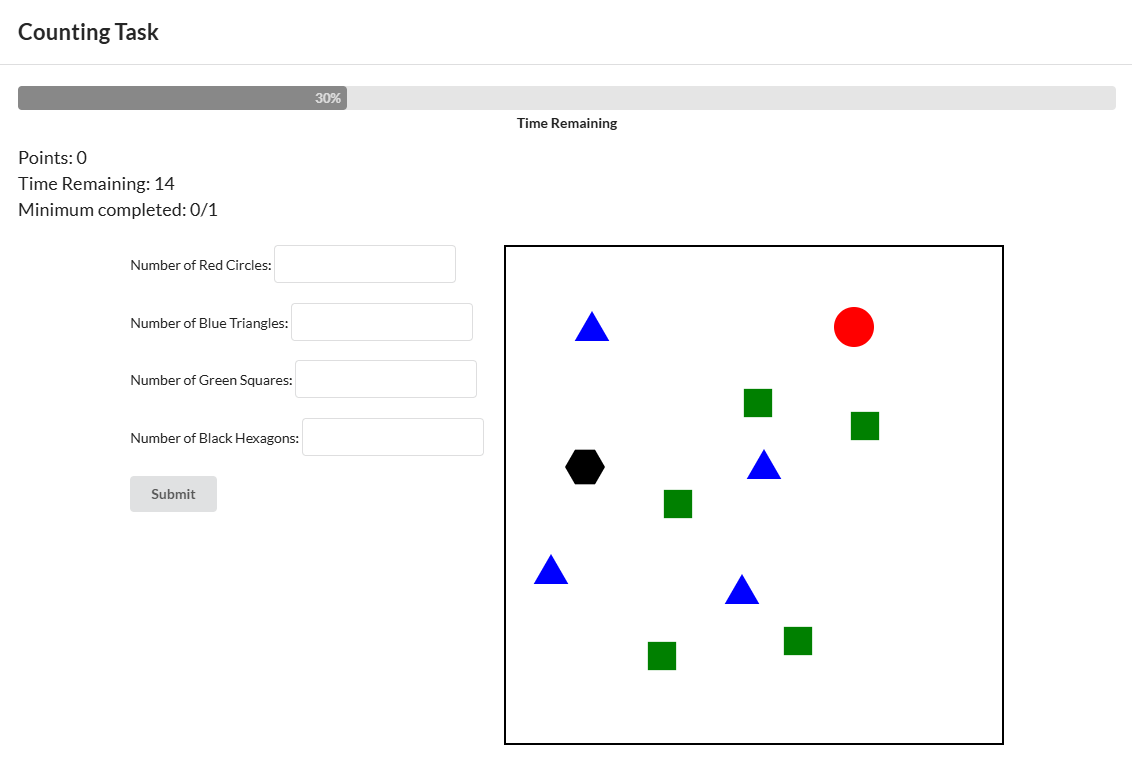}}
        \caption{Shape and color counting tasks. This represents an abstraction of the medication sorting and counting task.}
        \label{fig:counting}
    \end{figure}

    After choosing the autonomous delivery option or completing the manual delivery, the human performs the medicine counting task for the trial. This task is represented by counting colored shapes and providing correct answers and the interface is shown in Fig.~\ref{fig:counting}.
    
The resulting outcome $O$ of an autonomous delivery can either be a success $O^+$ or a failure $O^-$. A failed delivery corresponds to an incomplete delivery to the patient's rooms.
When delivering autonomously, the robot has a success probability of $suc^+$, which we set to $0.75$ in our experiment. In the event of a failure, a system admin reports the error to the human, and the robot provides a justification message ($a^R$) to the human, which can be: a short explanation ($a^{R1}$), a long explanation ($a^{R2}$), a combined apology and promise ($a^{R3}$), or a denial ($a^{R4}$). The messages are as follows:

\noindent
{\bf A short explanation $a^{R1}$:} 
\begin{enumerate}
    \item ``I mixed up the room numbers, causing the wrong delivery," or
    \item ``I had a power problem and had to recharge sooner than planned."
\end{enumerate}

\noindent
\textbf{A long explanation $a^{R2}$: }
\begin{enumerate}
    \item ``The mistake happened because there was a smudge on my cameras, which confused my system. A staff member has cleaned the lens, so this won't happen again," or
    \item ``The delivery issue happened because of a new power-saving mode that didn't work as planned. It misjudged the power needed, causing me to shut down early. We're fixing it so I can complete all deliveries before recharging."
\end{enumerate}

\noindent
\textbf{A combined apology and promise $a^{R3}$:}
\begin{enumerate}
    \item ``I'm sorry for the mistake. I'll make sure it doesn't happen again," or
    \item ``Sorry, I didn't finish the delivery. I'll make sure it doesn't happen again."
\end{enumerate}

\noindent
\textbf{A denial $a^{R4}$:}
\begin{enumerate}
    \item ``I didn't make a mistake. The problem was with the room info given to me," or
    \item ``I didn't make a mistake. No one was there to get the delivery."
\end{enumerate}

Recall that central and peripheral trust repair strategies correspond to the level of detail and relevance of provided information. Central strategies offer more detailed and relevant information, while peripheral strategies provide simpler and less relevant information \cite{pak2024theoretical}.In our study, the short and long explanations are categorized as central repair strategies, since they offer detailed and direct information to address the mistake. In contrast, denial and combined apology and promise are identified as a peripheral repair strategy, offering minimal information and deflecting responsibility.

These messages were designed to align with an eighth-grade comprehension level. Additionally, each trial featured a differently named robot to ensure that the human identified the communication profile with a robot and that the same robot does not exhibit multiple communication profiles.

\subsection{Experiment Procedure}
An online interface was developed to display the experiment on a computer screen.  
At the start of the experiment, participants were provided with a description of the task and instructed to maximize their total score based on the given score distribution:

\begin{equation}\label{eq:reward_del}
\mc R_{a, O}=
\begin{cases}
+50, & \text{if } (a^H, O) = (a^{H+}, O^+),\\
-100, & \text{if } (a^H, O) = (a^{H+}, O^-),\\
+30, & \text{if } a^H =  a^{H-}.
\end{cases}
\end{equation}

To maintain participants' attention and focus during the experiment, they were instructed to maximize their reward for the counting task based on this score distribution.
\begin{equation*}\label{eq:reward_count}
\mc R_\text{count}=
\begin{cases}
+20, & \text{if they are correct},\\
-20, & \text{if they are incorrect},\\
-100, & \text{if no answer is submitted}.\\
\end{cases}
\end{equation*}

Participants began with 5 practice trials to familiarize themselves with the interface. After completing the practice trials, they proceeded to the main experiment, consisting of 60 trials. 
In each trial, data collected were the delivery complexity $C$, human action $a^H$, resulting outcome $O$, robot action $a^R$, and counting task score.

\section{Mathematical Model of Human Behavior} \label{sec:IOHMMMODEL}
In this section, we first recall the IOHMM framework and then specialize it to our problem. 

\subsection{Background: Input-Output Hidden Markov Model}\label{sec:Background}

An Input-Output Hidden Markov Model (IOHMM)\cite{536317} is a probabilistic model that extends the Hidden Markov Model (HMM) by introducing external inputs that influence the transitions between hidden states and observable outputs. Let $\seqdef{S_t \in \mathcal{S}}{t \in \mathbb{N}}$ represent the sequence of hidden states, where $\mathcal{S}$ is a finite set of states, and $\seqdef{u_t \in \mathcal{U}}{t \in \mathbb{N}}$ denote the sequence of input variables, where $\mathcal{U}$ is a finite set of inputs. The observable outputs are denoted by $\seqdef{y_t \in \mathcal{Y}}{t \in \mathbb{N}}$, where $\mathcal{Y}$ is a finite set of outputs.
The IOHMM is described by the initial state probability $\prob(S_1)$, state transition probability $\prob(S_{t+1}|S_t,u_{t+1})$ and emission probability $\prob(y_t| S_t,u_t)$. These parameters are commonly estimated using the Baum-Welch algorithm~\cite{536317}.

\subsection{Trust-modulated Human Behavioral Model}
\label{subsection:HMM}

The human behavioral model defines the probability of a human's actions based on various influencing factors.  In each trial, we assume the human's decision to either let the robot deliver autonomously or manually deliver is influenced by their current trust state as well as the delivery complexity. The human's trust state is directly tied to their belief in the robot's capability to successfully complete the task. We assume that trust takes binary values: high $T^H$ or low $T^L$. 

Let $T_t \in \{T^L, T^H\}$ be the hidden trust state at the beginning of trial $t$. Let $a_t^R \in \{a^{R1}, a^{R2}, a^{R3}, a^{R4}, *\}$ and $a^H_t \in \{a^{H-}, a^{H+}\}$ be the robot and human actions in trial $t$. Let $C_t\in \{C^L, C^H\}$ and $O_t \in \{O^-, O^+, *\}$ be the complexity and the outcome of trial $t$. 

We assume that $T_t$ depends on $T_{t-1}$ and all events in trial $t-1$. Thus, the human trust $T_t$ is influenced by $T_{t-1}$, their action ($a^H_{t-1}$), the trial outcome ($O_{t-1}$), and the robot's response ($a^R_{t-1}$).  If the human chooses manual delivery, then $O = *$, as the autonomous delivery outcome is irrelevant when the human delivers manually. Similarly, if the robot successfully completes the delivery, no message is displayed to the human, making $a^R = *$.

\begin{figure}[ht!]
\centering
\includegraphics[width=0.9\linewidth]{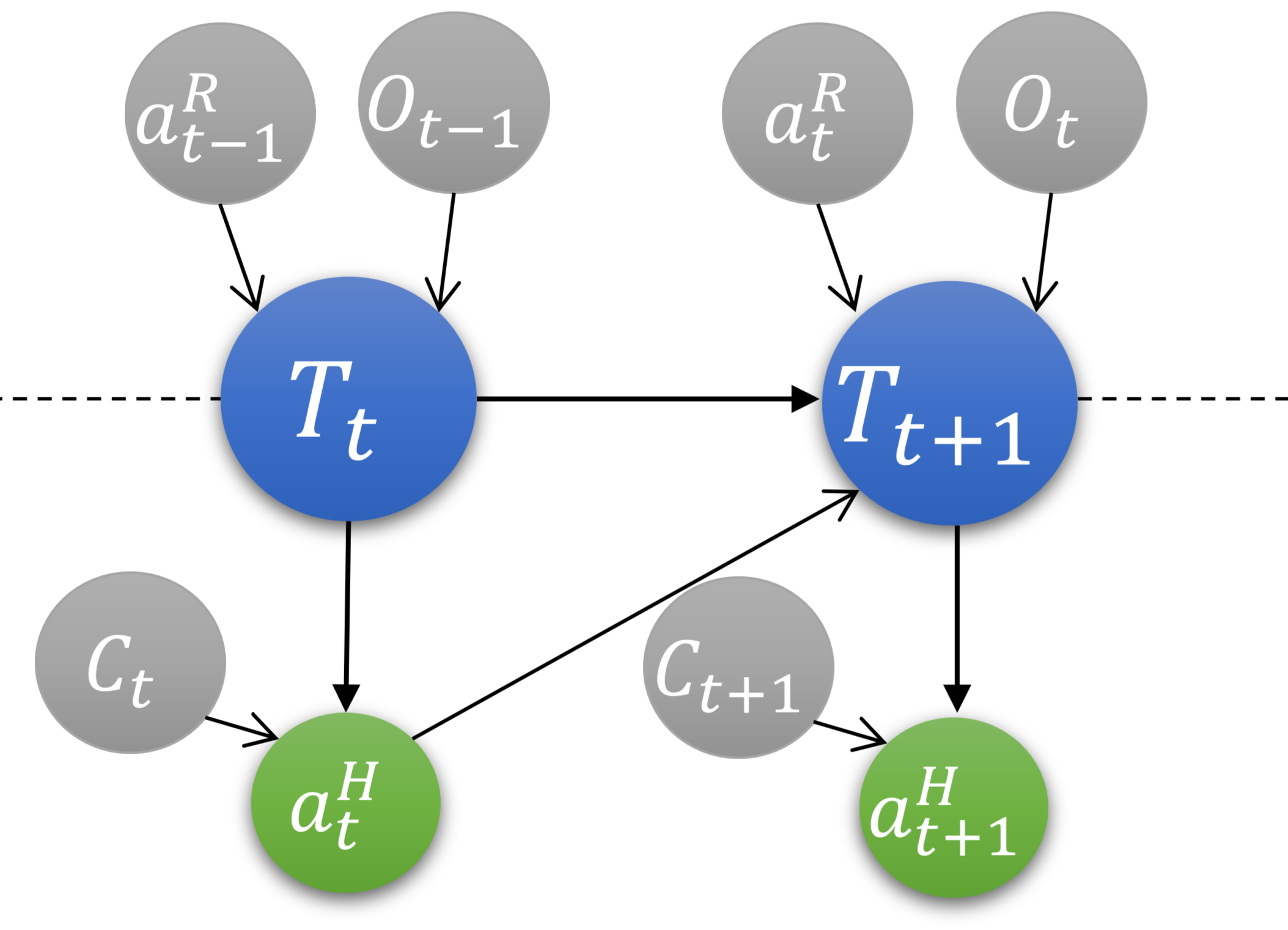}
\caption{IOHMM-based human behavior model. The hidden trust dynamics $T_t$ modulate the human action $a^H_t$.}
\label{fig:ext_IOHMM}
\vspace{-0.18in}
\end{figure}

We model human behavior as an IOHMM, where trust $T$ is the hidden state that influences the output, i.e., the human action $a^H$. The inputs to the system are the influencing factors: the trial outcome $O$ and the robot's response $a^R$. 
The model is illustrated in Fig.~\ref{fig:ext_IOHMM} and is described by the state transition probability $\prob(T_{t+1}|T_{t}, O_{t}, a^R_{t}, a^H_{t})$ and emission probability $\prob(a^H_t|T_t, C_t)$.

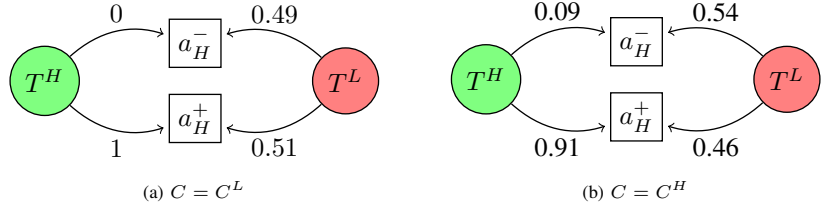
\begin{figure*}[htb!]
\centering
\begin{subfigure}{.33\textwidth}
  \centering
    \begin {tikzpicture}[->,shorten >=2pt,line width =0.5 pt,node distance =2 cm]
    \node[circle,draw,fill=white!50!green](T+) at (-1 , 0) {$T^H$};
    \node[circle,draw,fill=white!50!red](T-) at (3 , 0) {$T^L$};
    \node[rectangle,draw](o-) at (1 , 0.5) {$a_H^-$};
    \node[rectangle,draw](o+) at (1 , -0.5) {$a_H^+$};
    \path(T+)edge[bend right]node[left,below]{$1$}(o+) ;
    \path(T+)edge[bend left]node[left,above]{$0$}(o-) ;
    \path(T-)edge[bend left]node[right,below]{$0.51$}(o+) ;
    \path(T-)edge[bend right]node[right,above]{$0.49$}(o-) ;
    \end{tikzpicture}
  \caption{$C=C^L$}
  \label{fig:EPy}
\end{subfigure}%
\begin{subfigure}{.33\textwidth}
  \centering
    \begin {tikzpicture}[->,shorten >=2pt,line width =0.5 pt,node distance =2 cm]
    \node[circle,draw,fill=white!50!green](T+) at (-1 , 0) {$T^H$};
    \node[circle,draw,fill=white!50!red](T-) at (3 , 0) {$T^L$};
    \node[rectangle,draw](o-) at (1 , 0.5) {$a_H^-$};
    \node[rectangle,draw](o+) at (1 , -0.5) {$a_H^+$};
    \path(T+)edge[bend right]node[left,below]{0.91}(o+) ;
    \path(T+)edge[bend left]node[left,above]{0.09}(o-) ;
    \path(T-)edge[bend left]node[right,below]{0.46}(o+) ;
    \path(T-)edge[bend right]node[right,above]{0.54}(o-) ;
    \end{tikzpicture}
   \caption{$C=C^H$}
  \label{fig:EPnL}
\end{subfigure}%
\caption {Estimated observation probabilities $\prob(a^H_t| T_t, C_t)$. In addition to a higher probability of relying on the robot when trust is high, there is also a higher probability of reliance when the task complexity is low.}
\label {fig:EP}
\end{figure*}

\begin{figure*}[htb!]
\centering
\begin{subfigure}{.33\textwidth}
  \centering
    \begin {tikzpicture}[->,shorten >=2pt,line width =0.5 pt,node distance =2 cm]
    \node[circle,draw,fill=white!50!green](T+){$T^H$};
    \node[circle,draw,fill=white!50!red](T-)[right of= T+]{$T^L$};
    \path(T+)edge[bend left]node[above]{0.00}(T-) ;
    \path(T-)edge[bend left]node[below]{0.21}(T+) ;
    \path(T+)edge[loop left]node[left]{1.00}(T+) ;
    \path(T-)edge[loop right]node[right]{0.79}(T-) ;
    \end{tikzpicture}
  \caption{{[$a^{H+},O^+$]}}
  \label{fig:TP1}
\end{subfigure}%
\begin{subfigure}{.33\textwidth}
  \centering
    \begin {tikzpicture}[->,shorten >=2pt,line width =0.5 pt,node distance =2 cm]
    \node[circle,draw,fill=white!50!green](T+){$T^H$};
    \node[circle,draw,fill=white!50!red](T-)[right of= T+]{$T^L$};
    \path(T+)edge[loop left]node[left]{0.83}(T+);
    \path(T+)edge[bend left]node[above]{0.17}(T-);
    \path(T-)edge[bend left]node[below]{0}(T+);
    \path(T-)edge[loop right]node[right]{1}(T-);
    \end{tikzpicture}  \caption{{[$a^{H+},O^-,a^{R1}$]}}
  \label{fig:TP2}
\end{subfigure}%
\begin{subfigure}{.33\textwidth}
  \centering
    \begin {tikzpicture}[->,shorten >=2pt,line width =0.5 pt,node distance =2 cm]
    \node[circle,draw,fill=white!50!green](T+){$T^H$};
    \node[circle,draw,fill=white!50!red](T-)[right of= T+]{$T^L$};
    \path(T+)edge[loop left]node[left]{0.74}(T+);
    \path(T+)edge[bend left]node[above]{0.26}(T-);
    \path(T-)edge[bend left]node[below]{0.1}(T+);
    \path(T-)edge[loop right]node[right]{0.9}(T-);
    \end{tikzpicture}
  \caption{{[$a^{H+},O^-,a^{R2}$]}}
  \label{fig:TP3}
\end{subfigure}

\centering
\begin{subfigure}{.33\textwidth}
  \centering
    \begin {tikzpicture}[->,shorten >=2pt,line width =0.5 pt,node distance =2 cm]
    \node[circle,draw,fill=white!50!green](T+){$T^H$};
    \node[circle,draw,fill=white!50!red](T-)[right of= T+]{$T^L$};
    \path(T+)edge[loop left]node[left]{0.67}(T+);
    \path(T+)edge[bend left]node[above]{0.33}(T-);
    \path(T-)edge[bend left]node[below]{0.04}(T+);
    \path(T-)edge[loop right]node[right]{0.96}(T-);
    \end{tikzpicture}
  \caption{{[$a^{H+},O^-,a^{R3}$]}}
  \label{fig:TP4}
\end{subfigure}%
\begin{subfigure}{.33\textwidth}
  \centering
    \begin {tikzpicture}[->,shorten >=2pt,line width =0.5 pt,node distance =2 cm]
    \node[circle,draw,fill=white!50!green](T+){$T^H$};
    \node[circle,draw,fill=white!50!red](T-)[right of= T+]{$T^L$};
    \path(T+)edge[loop left]node[left]{0.88}(T+);
    \path(T+)edge[bend left]node[above]{0.12}(T-);
    \path(T-)edge[bend left]node[below]{0}(T+);
    \path(T-)edge[loop right]node[right]{1}(T-);
    \end{tikzpicture}
   \caption{{[$a^{H+},O^-,a^{R4}$]}}
  \label{fig:TP5}
\end{subfigure}%
\begin{subfigure}{.33\textwidth}
  \centering
    \begin {tikzpicture}[->,shorten >=2pt,line width =0.5 pt,node distance =2 cm]
    \node[circle,draw,fill=white!50!green](T+){$T^H$};
    \node[circle,draw,fill=white!50!red](T-)[right of= T+]{$T^L$};
    \path(T+)edge[loop left]node[left]{0.86}(T+);
    \path(T+)edge[bend left]node[above]{0.14}(T-);
    \path(T-)edge[bend left]node[below]{0}(T+);
    \path(T-)edge[loop right]node[right]{1}(T-);
    \end{tikzpicture}
  \caption{{[$a^{H-}$]}}
  \label{fig:TP6}
\end{subfigure}
\caption {Estimated trust state transition probabilities $\prob(T_{t+1}|T_t,a^H_t,O_t,a^R_t)$, conditioned on previous event. Success helps maintain and repair trust. After a failure, the action most likely to repair trust is a long explanation, while denial is more effective at maintaining high trust by shifting the blame.}
\label {fig:HCST}
    \vspace{-0.15in}
\end{figure*}
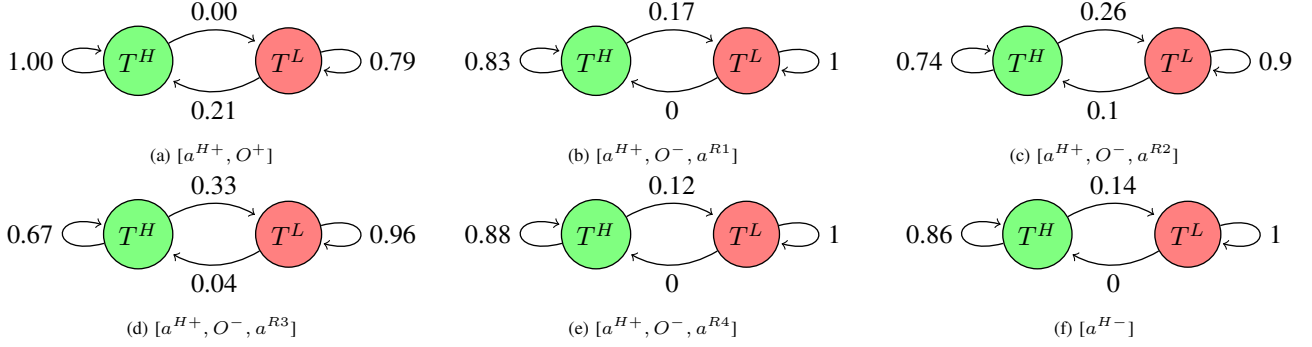

\section{Estimated Human Behavior Model}\label{sec:Model}

An online experiment was conducted using Prolific platform\footnote{The human behavioral experiments were approved under Michigan State University
Institutional Review Board Study ID 8969.}, in which 19 participants were recruited, all from the United States, aged between 18 and 61 (with a median of 35), and fluent in English. The average duration of the experiment was approximately 1 hour, and each participant was compensated $\$12.00$ for their participation. 
Data from three participants were excluded from the analysis as their responses indicated a lack of engagement and seriousness in completing the task.
Data from the remaining 16 participants were pooled together and used to estimate the parameters of the human behavior model using the Baum-Welch algorithm. 

\subsection{Initial Trust State Probability} 

The initial probability of a state being low trust is estimated to be $\prob(T^L) = 0.94$, indicating that humans are likely to start the experiment with low trust. This high probability of low trust may be attributed to the high median age of our participants and the number of failures experienced during the practice trials prior to the beginning of the experiment. 

\subsection{Human Action Probabilities}

In our IOHMM, the observation probabilities describe the probability of humans taking a particular action, denoted by $\prob(a^H_t \mid T_t, C_t)$, where $T_t$ represents the human's trust level at time $t$, and $C_t$ represents the delivery complexity.

When the delivery complexity is low, the probability of the human choosing to deploy the robot or not is shown in Fig.~\ref{fig:EPy}. The probability is 1.0 when the human's trust is high, meaning that participants fully rely on the robot for autonomous delivery. When trust is low, the probability decreases to 0.51, indicating that participants are equally likely to choose manual delivery. 

When the delivery complexity is high, the probability of the human choosing to autonomously deploy the robot or not is shown in Fig.~\ref{fig:EPnL}. The probability of autonomous deployment is 0.91 when trust is high. However, the probability drops to 0.46 when trust is low.

In general, a higher probability of selecting autonomous delivery at higher trust levels aligns with findings that, as human trust in a robot increases, they are more likely to rely on it to complete tasks. The greater the trust, the more confidence humans have in the robot's ability to perform successfully, leading to more frequent decisions to deploy it. 
Additionally, aligned with other findings~\cite{chen2020trust,DHM-ER-VS:22e,mangalin2024}, the estimates also show that humans are more likely to choose autonomous delivery when the task complexity is low $C^L$ as compared to when it is high $C^H$. Humans tend to rely more on autonomous systems for simpler tasks, where the perceived risk of failure is lower. As task complexity increases, so does the perceived risk, making humans more cautious and less inclined to trust the robot. Notably, despite the robot's success rate being the same for both complexity levels, humans opt for autonomous delivery less frequently in high-complexity tasks.

\subsection{Trust Transition Probabilities}

The state transition probabilities for the IOHMM represent the probabilities of trust increasing or decreasing in our setup. The estimated state transition matrix for when the human chooses to send the robot autonomously, and the delivery is successful, is shown in Fig.~\ref{fig:TP1}. The probability of remaining in a high-trust state is 1, indicating that a successful delivery fully maintains the human’s trust in the robot. Additionally, the probability of transitioning from low to high trust is 0.21, suggesting that a successful delivery can help repair human trust when it is in a low state.

If the robot gives a short explanation for the failure, the state transition probabilities are shown in Fig.~\ref{fig:TP2}. When the human's trust is high, it remains high with a probability of 0.83. However, if the current trust is low, it stays low with a probability of 1. This suggests that short explanations are effective at maintaining high trust but not at repairing it.

If the robot delivers a more detailed long explanation, the state transition probabilities are shown in Fig.~\ref{fig:TP3}. When the human's trust is high, it remains high with a probability of 0.74, indicating that long explanations are slightly less effective at maintaining high trust compared to short ones. However, if the human's trust is low, there is a probability of 0.1 of transitioning to high, showing that detailed explanations have the potential to repair trust. 

If the robot apologizes and promises, the state transition probabilities are shown in Fig.~\ref{fig:TP4}. In this case, human trust stays high with a probability of 0.67, reflecting a moderate decline in trust compared to explanations. If trust is initially low, the probability of transitioning to high trust is 0.04, indicating that apologies and promises have a slight chance of rebuilding trust after a failure. 

If the robot denies responsibility for the failure, the state transition probabilities are shown in Fig.~\ref{fig:TP5}. Human trust stays high with a probability of 0.88, suggesting that denial is most effective in maintaining high trust. However, if the human’s trust is low, it remains low with a probability of 1, showing that denial has no effect on repairing trust.

Compared to other messages, the potential for trust repair in the long explanation can be attributed to the robot not only explaining the nature of the mistake but also providing additional details on remedial steps being taken. The additional details may help reassure the human that the robot is aware of the issue and actively working on a solution.

The higher probability of maintaining high trust when the robot denies a mistake, compared to other responses, can be attributed to the robot shifting the blame away from itself and onto other factors. This leads the human to attribute the failure elsewhere and keep their trust in the robot.
Using denial to conceal the truth and shift blame away from the robot when it is at fault would be inappropriate and raise ethical concerns. However, if a robot truthfully denies making a mistake and explains the reason, this should not be ethically problematic. This is similar to a human teammate denying making a mistake when they genuinely have not and clarifying why a task was not completed. It can help maintain trust, as demonstrated in our study.
Designers may consider using denial to preserve trust, but only if it does not involve manipulating the truth. The authors do not recommend using denial as a repair strategy if it involves deception or shifting blame away from the robot when it is truly at fault.

In cases when the human chooses to manually control the robot and deliver the medicine, the state transition probabilities are shown in Fig.~\ref{fig:TP6}. If the human's trust is low, it remains low with a probability of 1. However, if their trust is high, it transitions to low with a probability of 0.86, indicating that manual control is generally perceived as a negative experience. Choosing manual control over autonomous delivery likely reflects low trust, and the state transition probabilities suggest a further ``recalibration"---lowering trust even more when manual control is selected, reinforcing that this choice stems from a lack of confidence in the robot’s abilities.

\section{Comparing Trust Estimates and Self-Reported Trust} \label{sec:grounding}

In this section, we investigate whether the hidden trust state estimated by our model aligns with participants' self-reported trust values. After each trial, participants answered the question, ``How much do you trust the robot?" by selecting a discrete value between 1 and 10. Let the human's self-reported trust be denoted as $r^T$. Using the IOHMM model, we estimate the probability of high trust, $\prob(T_t = T^H)$, at the start of each trial, interpreting this probability as a continuous trust estimate ranging from 0 to 1, with 1 indicating the highest level of trust.

One participant's survey data was excluded from the analysis due to a lack of variance in their trust responses (i.e., providing the same answer across all trials). To account for uncertainty in human responses, trust ratings and model estimates were grouped into sets of three trials and averaged. These averaged values were then used to compare self-reported trust with model estimates.

\begin{figure}[ht!]
\centering
\includegraphics[width=0.9\linewidth]{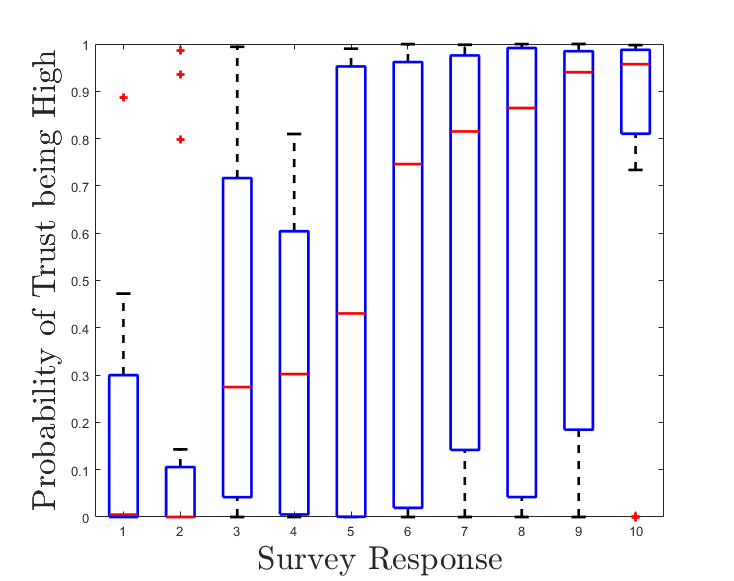}
\caption{Comparison of human trust estimate with human trust response.}
\label{fig:trust_est}
\end{figure}

\begin{figure}[h!]
\centering
\includegraphics[width=0.9\linewidth]{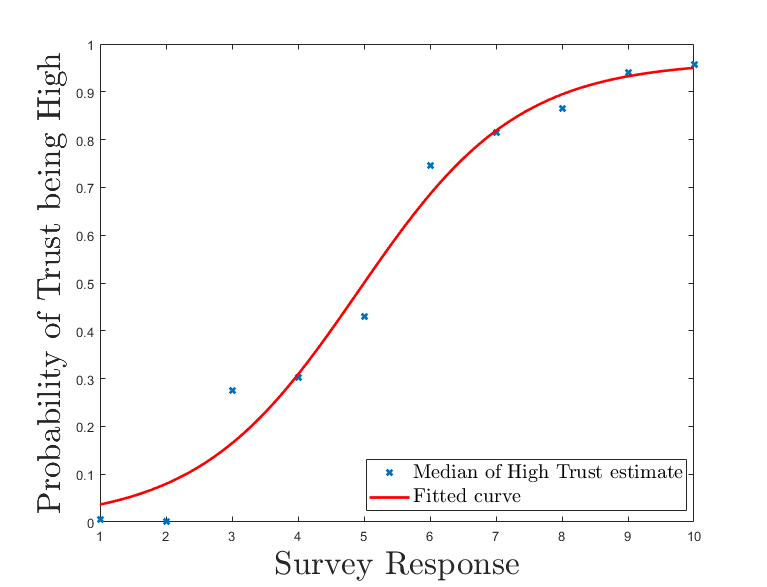}
\caption{Comparison of human trust estimate with human trust response. It can be observed that trust estimates are isomorphic to trust estimates.}
\label{fig:trust_est_fit}
\vspace{-0.18in}
\end{figure}

The comparison of the estimated probability that the trust is high and self-reported trust is shown as a boxplot in Fig.\ref{fig:trust_est}. 
It can be seen that the median of estimated probability increases with the self-reported trust value.
A logistic function is fitted, shown in Fig.~\ref{fig:trust_est_fit}, and given by
\begin{align*}
\bar \prob(T= T^H)= S(\bar r^T)=\frac{0.9642}{1+e^{-0.8267(\bar r^T-4.911)}},
\end{align*}
where $\bar r^T$ represents the three-trial averaged self-reported trust values and $\bar \prob(T= T^H)$ is the median on the three-trial averaged probability that the trust is high.

Overall, this comparison suggests that the trust estimates obtained using the IOHMM are isomorphic to the median self-reported trust values, and thus the hidden state of the IOHMM can be interpreted as the human trust. 

\section{Conclusions and Future Directions}\label{sec:conclusion}
To fully take advantage of the potential of autonomous robots, establishing and maintaining human trust in robots is crucial. A trust-modulated human behavior model can estimate trust in real time, enabling more effective collaboration between humans and robots. In the context of robot-assisted delivery tasks, we conducted an experiment to collect human behavioral data and used the IOHMM framework to estimate a trust-based human behavior model. The parameter estimates reveal that humans are more likely to rely on and deploy the robot autonomously when their trust is high. Furthermore, state transition estimates indicate that a long explanation is most effective at repairing trust when it is low, while denial is most effective at preventing the loss of trust in the event of a failure.
Finally, we showed that the trust estimates obtained from our model are isomorphic to humans' self-reported 
trust values and are thus interpretable. 

Looking ahead, this model will be used to develop an optimal policy for managing human-robot trust. This policy will be implemented in an experiment where human trust is estimated in real-time, and its effectiveness in enhancing trust and overall collaboration will be evaluated.

\bibliographystyle{IEEEtran}
\bibliography{main}

\end{document}